\documentclass[11pt]{article}

\usepackage{flafter}
\usepackage{jmlr2e}
\usepackage{amsmath}
\usepackage{graphicx}
\usepackage{subcaption}
\usepackage{listings}
\usepackage{algorithm}
\usepackage{algpseudocode}
\usepackage[table,xcdraw]{xcolor}

\usepackage{xcolor}

\ShortHeadings{DRL for  Contact-Rich Skills Using Compliant Movement Primitives}{Oren Spector and Miriam Zacksenhouse}
\firstpageno{1}

\begin{document}

\title{Deep Reinforcement Learning for Contact-Rich Skills Using Compliant Movement Primitives}

\author{\name Oren Spector \email oren.spector@campus.technion.ac.il \\
       \addr Faculty of Mechanical Engineering\\
       Technion– Israel Institute of Technology\\
       \AND
       \name  Miriam Zacksenhouse  \email mermz@me.technion.ac.il \\
       \addr Faculty of Mechanical Engineering\\
       Technion– Israel Institute of Technology\\
  }


\maketitle

\begin{center}
This work has been submitted to the IEEE for possible publication.  Copyright may be transferred without notice, after which this version may no longer be accessible. 
\end{center}

\begin{abstract}
In recent years, industrial robots have been installed in various industries to handle advanced manufacturing and high precision tasks. However, further integration of industrial robots is hampered by their limited flexibility, adaptability and decision making skills compared to human operators. 
Assembly tasks are especially challenging for robots since they are contact-rich and  sensitive to even small uncertainties. 
While reinforcement learning (RL) offers a promising framework to learn contact-rich control policies from scratch, its applicability to high-dimensional continuous state-action spaces remains rather
limited due to high brittleness and sample complexity. To address those issues, we propose different pruning methods that facilitate convergence and generalization. In particular, we divide the task into free and contact-rich sub-tasks, perform the control in Cartesian rather than joint space, and parameterize the control policy.  
Those pruning methods are naturally implemented within the framework of dynamic movement primitives (DMP). To handle contact-rich tasks, we extend the DMP framework  by introducing a coupling term that acts like the human wrist and provides  active compliance under contact with the environment. We demonstrate that the proposed method can learn insertion skills that are invariant to space, size, shape, and closely related scenarios, while handling large uncertainties. Finally we demonstrate that the learned policy can be easily transferred from simulations to real world and achieve  similar performance on UR5e robot. \\
\end{abstract}


\section{Introduction}
\label{s:INTRO}
The role of industrial robots in modern manufacturing has increased significantly in recent decades. Robots excel in tasks requiring repetitive movements with high precision, but are limited in their flexibility, ability to adapt to even small changes and decision making (\cite{michalos2014robo}). To overcome those limitations, the environment in which robots operate has to be arranged precisely and the robot operation has to be specified in details, resulting in high set up cost. Thus, industrial robots are mainly used in high volume manufacturing, when the significant set-up cost is justified ($25\%$ of large enterprises in the EU use robots compared to only  $5\%$ of small enterprises (\cite{EUstatistics}). Tasks that require flexibility and adaptability are usually designated to human operators. However, human operators lack the strength, endurance, speed, and accuracy of robots. Therefore, human operators create bottlenecks in production, resulting in higher costs, lower production rate and compromised product quality (\cite{boysen2008assembly}).

Assembly tasks are especially challenging for robots since they are contact-rich and  involve inherent uncertainties. Assembly tasks usually involve a sequence of two basic tasks: pick-and-place and peg-in-hole. Peg insertion may involve large parts, as in engine assembly (\cite{su2012new,su2012sensor}), or small electronic (\cite{lin2018development}) and even micro-products (\cite{Chang2011}). Thus, peg insertion is a common task in a wide range of industries, but the task details, including size and shape, may differ considerably.

While it may look simple, peg insertion is a very delicate action since even small tracking or localization errors may result in collisions. Tracking errors are expected to become larger as production rates increase due to inertia and low stiffness of serial robots.   Visual localization  is subject to errors, which depend on different parameters including image resolution, illumination, camera parameters and relative pose and location. Tracking and localization errors are inherent in assembly processes, so a good policy should be able to handle large errors to facilitate flexibility and speed.

Peg-in-hole has been extensively investigated over the years (starting with \cite{lozano1984automatic}, \cite{bruyninckx1995peg}). Robotic peg-in-hole usually involve two main phases: searching and inserting. During searching, the holes are identified and localized to provide the essential information required for inserting the pegs. Searching may be based on vision (\cite{Chang2011}, \cite{Wang2008}) or on blind  strategies involving, for example, spiral paths (\cite{Chhatpar2001}). Visual techniques depend greatly on the location of the camera, board and obstructions, and are about 3 times slower than human operators. Due to the limitations of visual methods, most strategies rely on force-torque and haptic feedback, either exclusively or in combination with vision. Combining vision and force control was demonstrated to facilitate generalization between cylindrical and cuboid pegs under large errors (\cite{Kim2014}), thought they did not show generalization over board location or unseen peg types. 

Insertion may involve contact model-based control or contact model-free learning (\cite{Xu2019}). Model-based strategies (\cite{Newman2001}, \cite{Jasim2014}, \cite{jasim2017contact}) estimate the state of the assembly from the measured forces, torques and position, and correct the movement accordingly using state-dependent pre-programmed compliant control. However, those strategies are limited to simple parts and do not generalize well. 

Model-free learning involves either learning from demonstration (LfD) or learning from the environment (LfE) (\cite{Xu2019}). LfD algorithms derive a policy from a set of examples or demonstrations provided by expert operators (\cite{Argall2009}). The flexibility of the resulting policy is limited by the information provided in the demonstration data set. 

Alternatively, the robot may learn the control policy by interacting with the environment and modifying the control policy using reinforcement learning (RL) (\cite{kober2013reinforcement}). The application of RL methods to manipulation tasks is challenging due to the  large amount of experience needed to learn successful policies. Mode-based RL methods can provide good sample efficiency, but require learning a good model of the dynamics. This can be achieved by learning prior knowledge of the system dynamics from previous tasks and adapting the model online to locally compensate for un-modeled variations (\cite{fu2016one}).  In contrast, model-free RL methods either learn the policy that maximizes the accumulated reward during new trajectories (policy-based methods) or learn a value function and derive the policy that maximizes the expected value at each step (value-based methods) (\cite{kober2013reinforcement},  \cite{xu2018feedback}). Value-based methods are limited due to brittle convergence, especially in continuous spaces, while policy-based methods are limited due to sample-inefficiency, since new trajectories have to be generated for each gradient step (\cite{kober2009learning}, \cite{peters2008reinforcement}).

Despite recent advances, there is still a large gap in flexibility and adaptability between robots and humans. In order to bridge this gap, we focus on acquiring general relevant skills rather than planning paths for specific problems. In contrast to planners,  skills should be able to generalize, at least partially, across: (1) space (hole location), (2) sizes, (3) shapes, and (4) closely related scenarios (e.g., inserting pegs with previously un-encountered shapes), and to handle large uncertainties. In particular, we focus here on insertion skills. 

The rest of the paper is structured as follow: Section \ref{GC} describes general considerations that motivated the selection of a state-of-the art model-free on-policy RL algorithm for contact-rich assembly tasks, and how we address the challenge of sample efficiency. Section \ref{DMP} presents the framework of dynamic movement primitives (\cite{ijspeert2002movement}) and introduces a novel  coupling term that provides active compliance. Section \ref{SimMethods} describes the simulation methods. Section \ref{Results_simulaion}  presents simulation results and demonstrates skill acquisition, i.e.,  handling large uncertainties and generalizing across space, sizes, and shapes including a previously un-encountered shape. Section \ref{s:Robot} demonstrates that the learned controller overcomes the sim-to-real challenge and performs successfully on a real robot (Universal Robots, UR5e). Finally, Section \ref{conclusion} discusses the merits of the proposed approach and outlines future work.

\section{General considerations}
\label{GC}
\subsection{End-to-end learning vs modular learning} 
End-to-end learning has been advocated lately as a generic approach to train a single learning system without breaking it down into  pre-conceived modules (\cite{schmidhuber2015deep}, \cite{collobert2011natural} , \cite{krizhevsky2012imagenet}). Thus, it replaces the modular approach to learning, in which feature extraction and representation were derived or learned separately from the policy or classification module. End-to-end learning is based on the assumption that more relevant features and better representations can be learned (and shared) in a single deep neural network (NN). 

However, a  recent study challenges the power of end-to-end learning (\cite{Glasmachers2017}). In particular, it was demonstrated that as networks grow, end-to-end learning can become very inefficient and even fail. Thus, structured training of separate modules may be more robust. The modular approach is especially suitable for assembly tasks, which  are naturally divided into different sub-tasks that can be learned in different modules. Imitation learning, for example, can facilitate learning the controller, while prior knowledge in the form of computer-aided design (CAD) can facilitate learning pose estimation.  Given those considerations, we adopt the modular approach, with two main modules (i) Visual processing and localization, and (ii) trajectory planning and control. Nevertheless, additional research is needed to address the advantages of end-to-end learning versus modular (or other) learning methods.

\subsection{Reinforcement Learning}
Reinforcement learning (RL)  encompasses two major methods: model-based and model-free RL algorithms. The two methods are  distinguished by the objective of the learning phase. The goal of model-based RL is to learn an approximated model, which is then used to derive an optimal controller $\pi$ that maximizes the reward function $R$. Model-based RL algorithms have shown some great success in robotic applications including robotic manipulation (\cite{fu2016one}, \cite{luo2018deep}). However, model based RL algorithms have an inherent disadvantage in contact-rich tasks, since they are non-linear and non-repeatable and thus difficult to model. The magnitude of friction, for example, is a factor of the surface roughness, which is not observable in most cases. Model-free algorithms are much more suitable for contact-rich tasks because they can  generalize to unknown parameters in the model, such as friction. Holding a glass, for example, involves a highly complicated model, though the policy itself is relatively easy to learn. 

Here we use model-free deep RL, where a NN is used for high-capacity function approximation to provide generalization. 
Model-free deep RL algorithms have been applied in a range of challenging domains, from games (\cite{mnih2013playing}, \cite{silver2016mastering}) to robotic control (\cite{haarnoja2018soft}, \cite{peters2008reinforcement}). 
Implementing those methods in real-world domains has two major challenges: (1) sample inefficiency, and (2) convergence and brittleness with respect to their hyper-parameters: learning rates and exploration constants. 

Model free RL can be divided to on-policy and off-policy methods. Sample inefficiency is especially critical in on-policy methods such as trust region policy optimization (TRPO, \cite{Schulman2015}), proximal policy optimization (PPO, \cite{Schulman2017})  and actor-critic (A3C, \cite{Mnih2016}), since new samples have to be collected for each gradient step. This quickly becomes extravagantly expensive, as steps needed to learn an effective policy increases with task complexity. Off-policy algorithms aim to reuse past experience, but tend to be brittle and are not guaranteed to converge in continuous state and action spaces (\cite{bhatnagar2009convergent}).  A commonly used algorithm in such settings, deep deterministic policy gradient (DDPG) (\cite{lillicrap2015continuous}), improves sample-efficiency  but is  extremely brittle. Here we choose the model-free on-policy algorithm PPO  since it is guaranteed to converge and is relatively less  brittle, and address sample inefficiency by using different pruning methods as detailed next.

\subsection{Cartesian versus joint space}
\label{ss:AS} 
The choice of action space has a strong impact on robustness and task performance as well as learning efficiency and exploration (\cite{Martin-Martin2019}). A common approach is to specify the actions in the joint space of the robot, so the policy determines the torques to be applied at the joints (\cite{levine2016end}). However, interaction forces are related to joint torques via the Jacobian of the robot, which depends non-linearly on the joint angles. Hence, a simple policy in the Cartesian space, such as a policy that applies a constant force, would require  learning a very complex NN to approximate the Jacobian at every pose. Additionally, errors in modeling the Jacobian would further impede the transfer of learning from simulations to real robots (\cite{peng2018sim}, \cite{chebotar2019closing}).
In contrast, position or force commands in the Cartesian space of the EEF can be sent to the internal controller of the robot, which uses the internally coded Jacobian to determine the joint-torques.  Thus, specifying the actions in the end effector (EEF) space can improve robustness and accelerate learning rate dramatically, especially in contact-rich environments.

\subsection{Reinforcement learning for contact rich tasks}

Following the previous considerations, we choose the model-free on-policy algorithm PPO  since it is guaranteed to converge and is relatively less  brittle. Sample-efficiency, which is the key disadvantage of on-policy algorithms, is addressed by using three methods. 
First we  work in Cartesian action space, which simplifies the learning problem especially when contact is involved,  as  explained in Section \ref{ss:AS}. 
Second, the task is naturally split into free movements and interaction movements, by constructing a reactive planner that becomes active only under external forces. Free movements have well-known solutions and in particular can be performed using proportional-differential (PD) controllers. Thus, we focus on learning the reactive controller.  This approach can be extended to more complex tasks by building on previously learned controllers. 
Third, the reactive controller is formulated to mimic the gentle trajectory corrections provided by the wrist and fingers, by introducing active compliance control, which is critical for contact-rich tasks (\cite{cutkosky2012robotic}, \cite{schumacher2019introductory}). Significant dimension reduction is obtained by learning just the parameters of the compliance and PD controllers. 

\section{Dynamic Movement Primitives  for compliant control}
\label{DMP}
The Dynamic Movement Primitive (DMP) framework (\cite{ijspeert2002movement}, \cite{schaal2006dynamic}, \cite{ijspeert2013dynamical})
addresses some of the issues raised in section \ref{GC}. First, it defines the movement in Cartesian space, rather than joint space, which is critical for contact-rich tasks. Most importantly, DMP  facilitates the integration of  different movement primitives (\cite{pastor2009learning}, \cite{hoffmann2009biologically}), which can be used to divide the task into known and unknown control policies for dimensional reduction. 
\subsection{Basic formulation} 
The main idea of DMP is to use dynamic systems to generate discrete or rhythmic  movements that converge to a desired point of attraction or a desired  limit cycle, respectively. Specifically, a discrete movement  $y(t)$ is generated by a non-linear dynamical system, which can be interpreted as a linear damped spring perturbed by an external non-linear forcing term $f$:
\begin{equation}
\begin{aligned}
\tau \ddot{y}=K_y(g-y)-D_y\dot{y}+(g-{{y}_{0}})f\
\end{aligned}
\label{eq:DMP}
\end{equation}
where g is a known goal state, $K_y$ and $D_y$ are spring and damping diagonal matrices, respectively, and $\tau$ is a scaling factor. In our case, $y$ is a six-dimensional vector $y=[x \ \ \theta]'$ that includes the 3-dimensional position $x$ and 3-dimensional orientation $\theta$ of the EEF in Cartesian space. 
For appropriate parameter settings and $f = 0$,  Eq. \ref{eq:DMP}  forms a globally stable linear dynamic system with $g$ as a unique point attractor.   The function $f$ can be used to modify the exponential convergence toward $g$ to allow complex trajectories. 

It is advantageous to formulate $f$ as a function of a phase variable, $x$, rather than time, to facilitate perturbation rejection, and control movement duration (\cite{ijspeert2002movement}).  The dynamics of the phase variable are given by:
\begin{equation}
\tau \dot{s}=-{{\alpha }_{s}}s
\end{equation}
The phase variable is initialized to $1$ and converges to zero. 

The non-linear function $f$ can be expressed as:
\begin{equation}
f(s)=\frac{\sum\limits_{i=1}^{N}{{{\psi }_{i}}{{\omega }_{i}}}}{\sum\limits_{i=1}^{N}{{{\psi }_{i}}}}s \\ 
\label{eq:DMP_2}
\end{equation}
with Gaussian basis functions
\begin{equation}
{{\psi }_{i}}=\exp \left( -\frac{1}{2{{\sigma_i }^{2}}}{{\left( s-{{c}_{i}} \right)}^{2}} \right)
\end{equation}
with parameters $c_i$ and $\sigma_i$. The weights $\omega_i$ can be learned to generate desired trajectories, using locally weighted regression. Thus, $f$ can be designed to manage the known part of the task, and, in particular, is designed here to generate minimum jerk trajectories in free space. The coupling term, considered next, is designed to modify those trajectories in response to interaction forces. 

\subsection{DMPs for compliant control} 
Humans perform contact-rish assembly tasks using their fingers and wrist for fine manipulations while the arm serve to perform  gross motions, as  in writing (\cite{cutkosky2012robotic}). Wrist movements may be  generated using  passive or active compliance control (\cite{Wolffenbuttel1990}). Active compliance may be slower than passive compliance but may generalize over different tasks (\cite{Wang1998}). 
A general way to describe active compliance is by a spring and damper system in Cartesian space: 
\begin{equation}
F=K_{c}\Delta{y} + D_{c}\Delta{\dot{y}}
\label{eq:compliance}
\end{equation}
where $F$ is a six-dimensional vector that includes the 3-dimensional forces and 3-dimensional moments, and $K_{c}$ and $D_{c}$ are  $6 \times 6$ matrices of  compliant stiffness and damping. The $6 \times 6$ matrices  $K_{c}$ and $D_{c}$ are not restricted to be diagonal nor block-diagonal, so they can provide generalized compliance. In particular, off-diagonal terms can couple deviations in one direction with forces or moments in another direction as well as couple angular deviations with forces, or transitional deviations with  moments. The deviation vector $\Delta{y}$ is the difference between the force-free desired position $y$ and the modified desired position that satisfies the active compliance. 

Given the small velocities expected in peg-in-hole tasks, we focus on stiffness control with $D_c=0$, so Eq. (\ref{eq:compliance}) can be inverted directly to compute the deviations:
\begin{equation}
\Delta{y} =  K_c ^{-1}F
\label{eq:Deltay}
\end{equation}

The original DMP is now re-written in terms of the modified desired trajectory rather than the force-free desired trajectory:
\begin{equation}
\begin{aligned}
\tau \ddot{y}=K_y(g-y+\Delta{y})-D_y\dot{y}+(g-{{y}_{0}})f(s)\
\end{aligned}
\label{eq:DMPm2}
\end{equation}

Given Eq (\ref{eq:Deltay}), Eq (\ref{eq:DMPm2}) has the form of Eq (\ref{eq:DMP}) with an additional coupling term (\cite{ijspeert2013dynamical}): 
\begin{equation}
\begin{aligned}
\tau \ddot{y}=K_y(g-y)-D_y\dot{y}+(g-{{y}_{0}})f(s) + \boldsymbol{c_c}  \
\end{aligned}
\label{eq:DMPwCC}
\end{equation}
where the coupling term provides compliance according to: 
\begin{equation}
{{c}_{c}}=K_y\Delta{y}=K_y K_c ^{-1}F
\label{eq:CC}
\end{equation}

\subsection{Learning DMPs for compliant control} 
\label{ss:LrnDMP}
The diagonal  spring matrix ($K_y$), and the full compliant stiffness matrix ($K_c$) are learned using PPO.  The diagonal spring matrix is restricted to be positive definite and to have the same constants on the upper and lower parts $K_y = diag[k_t, k_t, k_t, k_{\theta}, k_{\theta}, k_{\theta}]$,  where $k_t$ and $k_{\theta}$ are the translation and rotation stiffness constants. The diagonal damping matrix $D_y$ is determined from the diagonal spring matrix to achieve critical damping for both translation and rotation. This assures that the settling time is shortest. 
Thus, a NN was trained to determine the $38$ parameters of the PD ($k_t$ and $k_{\theta}$) and compliant ($K_c$) controllers given the hole location and shape.

\section{Simulation Methods}
\label{SimMethods}
\subsection{DMP implementation}
Eq (\ref{eq:DMPwCC}) can be  viewed as a PD controller in Cartesian space with a modified goal set by the  external nonlinear forcing term $f(s)$ and coupling term $c_c$:
\begin{equation}
\begin{aligned}
\tau \ddot{y}=K_y(g_m-y)-D_y\dot{y}\
\end{aligned}
\label{eq:DMPwCCmG}
\end{equation}
where, using Eq  (\ref{eq:CC}), the modified goal is given by:
\begin{equation}
\begin{aligned}
g_m=g+K_y^{-1}(g-{{y}_{0}})f(s) + \Delta y\
\end{aligned}
\label{eq:mG}
\end{equation}

Following Eq (\ref{eq:DMPwCCmG}), we force the robot to move at the required EEF acceleration, by applying the required joint torques. The dynamics of a rigid  robotic manipulator with $n$ degrees of freedom can be described by Euler-Lagrange equations in the joint space (\cite{Rodd1987}):
\begin{equation}
H(q)\ddot{q}+h(q,\dot{q})=u
\label{eq:DynJoint}
\end{equation}
where $q$ is the joint displacement vector, $H(q)$ is the $n \times n$ symmetric, positive definite inertia matrix, $h(q,\dot{q})$ is the $n \times 1$ vector of the joint torques due to centrifugal and gravity forces, and $u$ is the $n \times 1$ vector of the joint torques supplied by the actuators. Using the kinematic relationships $\dot y = J \dot q$, where $J$ is the Jacobian of the manipulator, and $\ddot y = \dot J \dot q + J \ddot q$, Eq (\ref{eq:DynJoint})   can be expressed in terms of the Cartesian acceleration  $\ddot{y}$ as (\cite{Valency2003}):
\begin{equation}
H^{*}(q)\ddot{y}+h^{*}(q,\dot{q})=u\\
\label{eq:DynCS}
\end{equation}
where $H^{*}=H(q)J^{+}$, $J^{+}$ is the pseudo inverse of $J$, and $h^{*}(q,\dot{q})=h(q,\dot{q})-H(q)J^{+}\dot{J}\dot{q}$.

Inserting Eq  (\ref{eq:DMPwCCmG}) in Eq (\ref{eq:DynCS}), the control signal defining the joint torques is given by:
\begin{equation}
u=H^{*} \tau^{-1} \large ( K_y(g_m - y)-D_y \dot y \large )+h^*
\end{equation}
This controller is asymptotic stable (\cite{hsia1990robot}) as long as $K_y$ and $D_y$ are positive definite matrices. As mentioned in Section \ref{ss:LrnDMP}, $K_y$ is restricted to be positive definite during training, and $D_y$ is derived from $K_y$ to achieve critical damping, thus assuring not only that it is  positive definite, but also that the settling time is shortest.  

\subsection{Task setup}
\label{TaskSU}
\begin{figure}[h!]
	\centering
	\begin{subfigure}[b]{0.44\linewidth}
		\includegraphics[height=3.8cm]{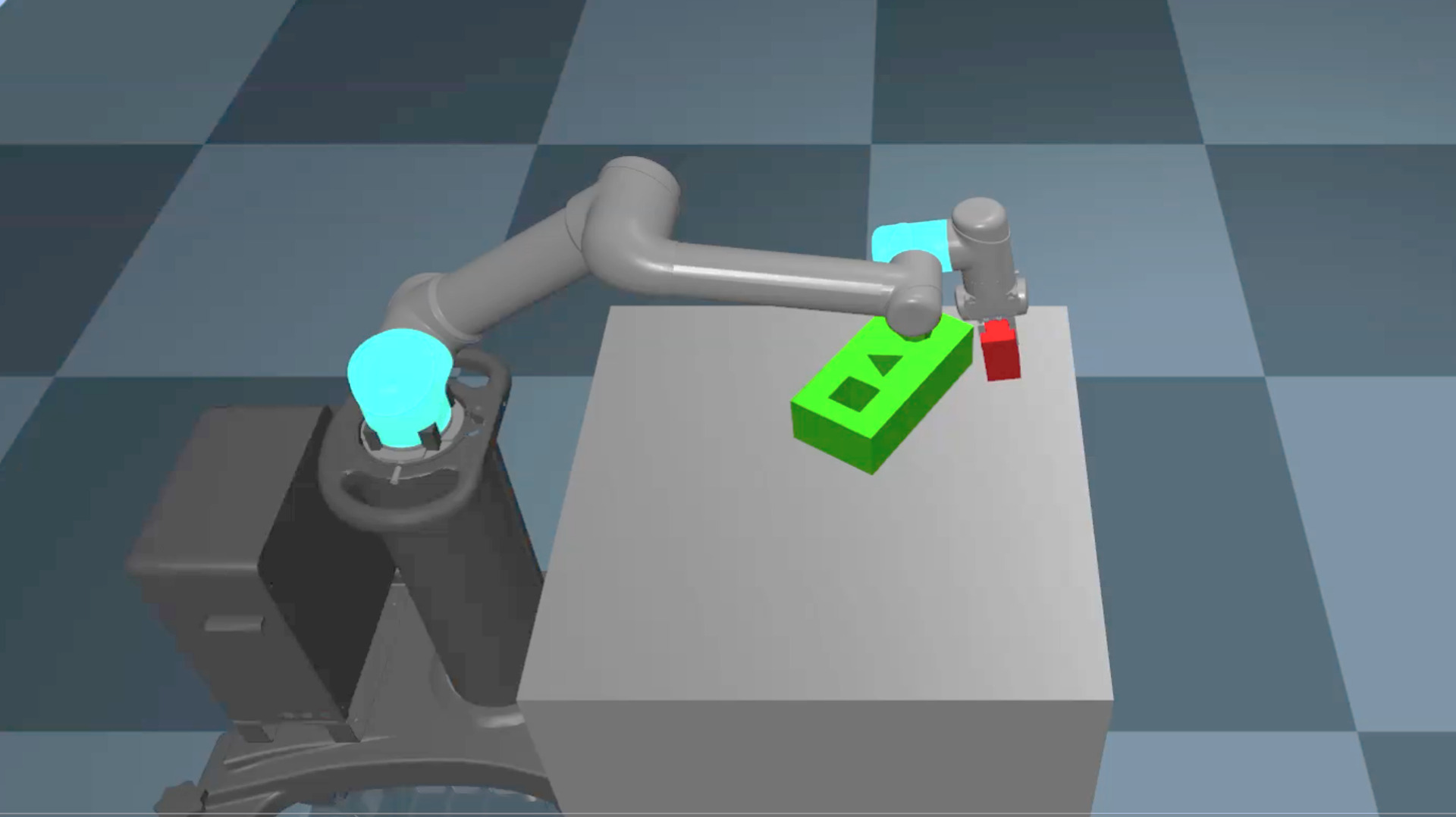}
		\caption{Peg type, Cube}
	\end{subfigure}
	\begin{subfigure}[b]{0.44\linewidth}
		\includegraphics[height=3.8cm]{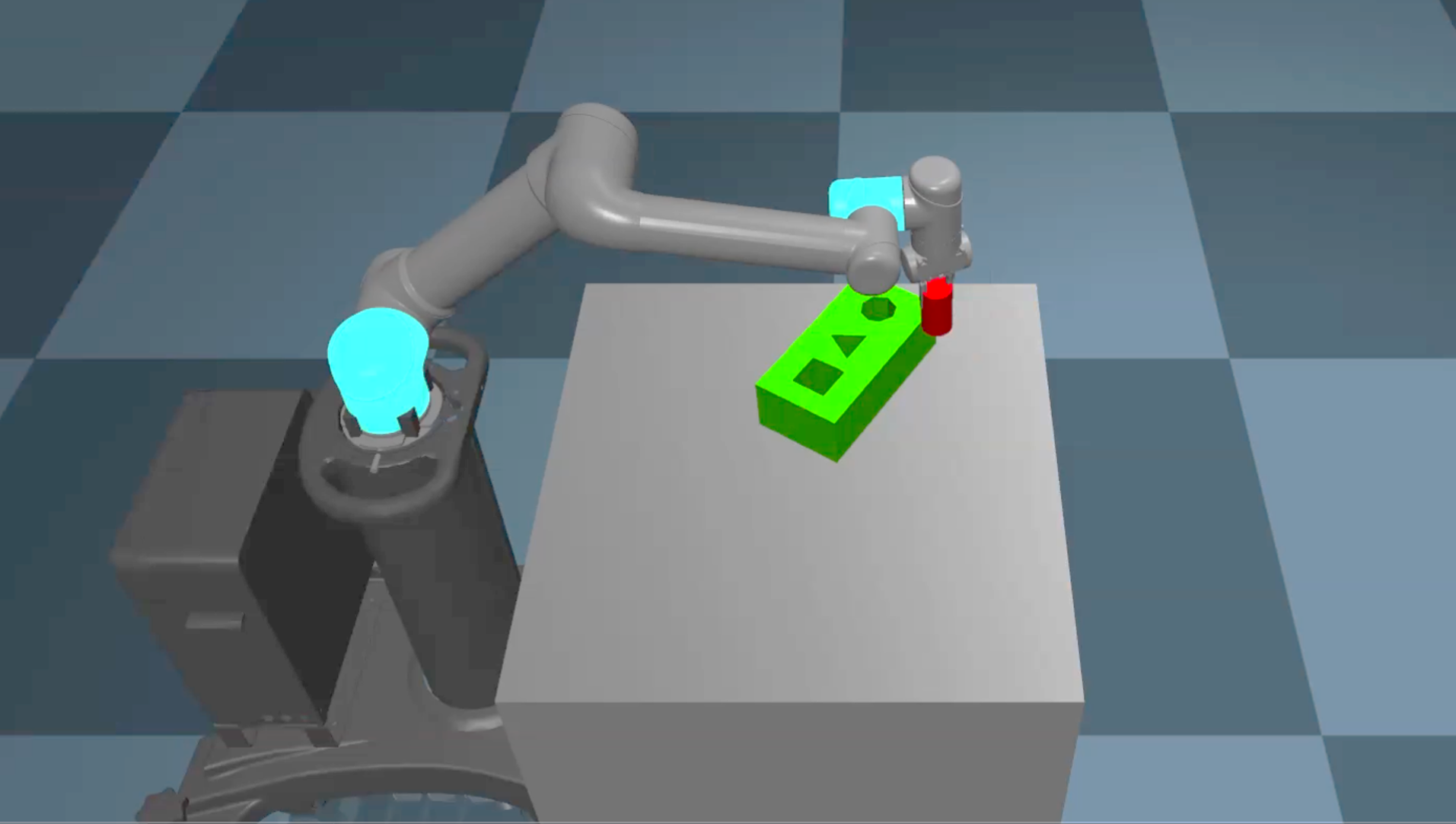}
		\caption{Peg type, Cylinder}
	\end{subfigure}
	\caption{Task setup.}
	\label{fig:Task_setup}
\end{figure}

\begin{figure}
	\centering
	\includegraphics[height=8cm]{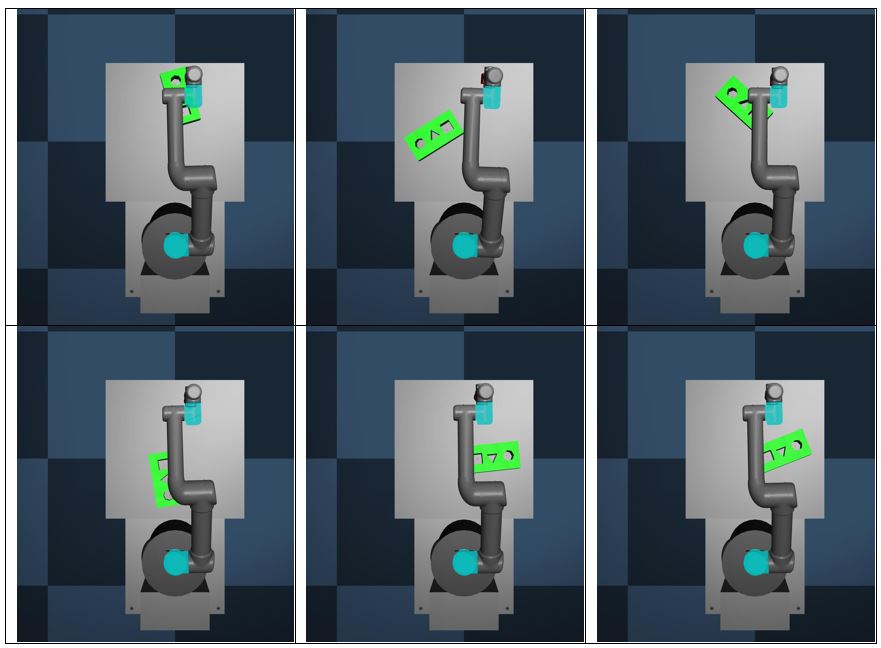}
	\caption{Simulated robot and environment with boards at different locations and angles.}
	\label{fig:board_location}
\end{figure}

The task was designed using RoboSuite (\cite{corl2018surreal}) to facilitate acquisition of generic skills for peg insertion,  as defined in Section \ref{s:INTRO}. To provide generalization over shapes, the task involves inserting pegs of different shapes into the corresponding holes in a board, as depicted in Figure \ref{fig:Task_setup}. The board is located in the $(x,y)$ plane  and the holes are along the $z$-axis.  Initial training is performed with cylindrical and cubic pegs, to demonstrate that a single policy can generalize over shape.  Testing and further training includes also pegs with triangular cross-section, to test generalization to closely related scenarios. Generalization over space is obtained by placing the board  in different locations in the workspace (relative to the robot, Figure \ref{fig:board_location}). Since the compliance is specified in the task space, we expect the parameters to be location-independent. Nevertheless we added location parameters to assist convergence (\cite{Dauphin2014}) and to present a more general framework. Finally we demonstrated also generalization to new sizes, both larger and smaller than  during training. 


Thus, training and testing was conducted under the following environmental conditions:
\begin{itemize}
	\item Physical/Geometrical parameters
	\begin{itemize}
		\item Orientation: the orientation of the board was uniformly distributed between $0-90^{\circ}$. 
		\item Location: the location of the board was uniformly distributed within the workspace (avoiding singularities). 
		\item Shape:  the cross section of the peg and corresponding hole was a circle (with diameter $D$), square or isosceles triangle (with edges of length $L$), so the policy was not trained for a specific shape.     
		\item Size: the characteristic size of the pegs $S=L=D$ was $S=50[mm]$ during training and $S=25,50,60$[mm] in testing. The characteristic size of the holes was $1.03D$ for circular holes and $1.08L$ for holes with square and triangular cross-sections. In particular, the characteristic size of the holes during training was $51.5$[mm] for circular holes and $54$[mm] for square holes. The depth of the holes was $d=80[mm]$. 
		\item Friction: the friction between the peg and the board was uniformly distributed between $0.2-0.9$ corresponding to steel/steel friction coefficient.
	\end{itemize}
\end{itemize}

In order to train a general controller, which can handle large uncertainties in localization and grasping, we inserted errors in the board location and pose:
\begin{itemize}
	\item Localization and grasping uncertainties  
	\begin{itemize}
		\item Visual localization uncertainties:  \textbf{Translational errors} in (x,y) were distributed uniformly within a range specified as percent of half the characteristic size.   \textbf{Orientation errors} around $z$-axis were uniformly distributed between $0-12^{\circ}$. 
		
		\item Grasping uncertainties: \textbf{Transnational errors} in $z$-axis reflect uncertainties in grasping height along the peg and were distributed as the translational localization errors.  \textbf{Orientation errors} around $(x,y)$-axes reflect  uncertainties in grasping angle and were distributed as the orientation errors due to visual localization. 
		
	\end{itemize}
\end{itemize}
The visual localization assumptions were validated by training a YOLO model based on a built-in 2-dimensional camera in MuJoCo.

\subsection{Reward}
Reward shaping is a very delicate craft, especially for general skills. We  divided the reward function to 2 parts: (1) Outcome reward given at the end of successful trials, and (2) shaped reward given during unsuccessful trials. A trial is considered successful when the tip of the peg is within a small distance from the table. The outcome reward is a constant of 38000 points. The shaped reward grants points relative to the distance between the tip of the peg and the desired destination, and is included to facilitate convergence. 

\section{Simulation Results} 
\label{Results_simulaion}
\subsection{Training curve}
\begin{figure}[h!]
	\includegraphics[height=9cm,width=15cm]{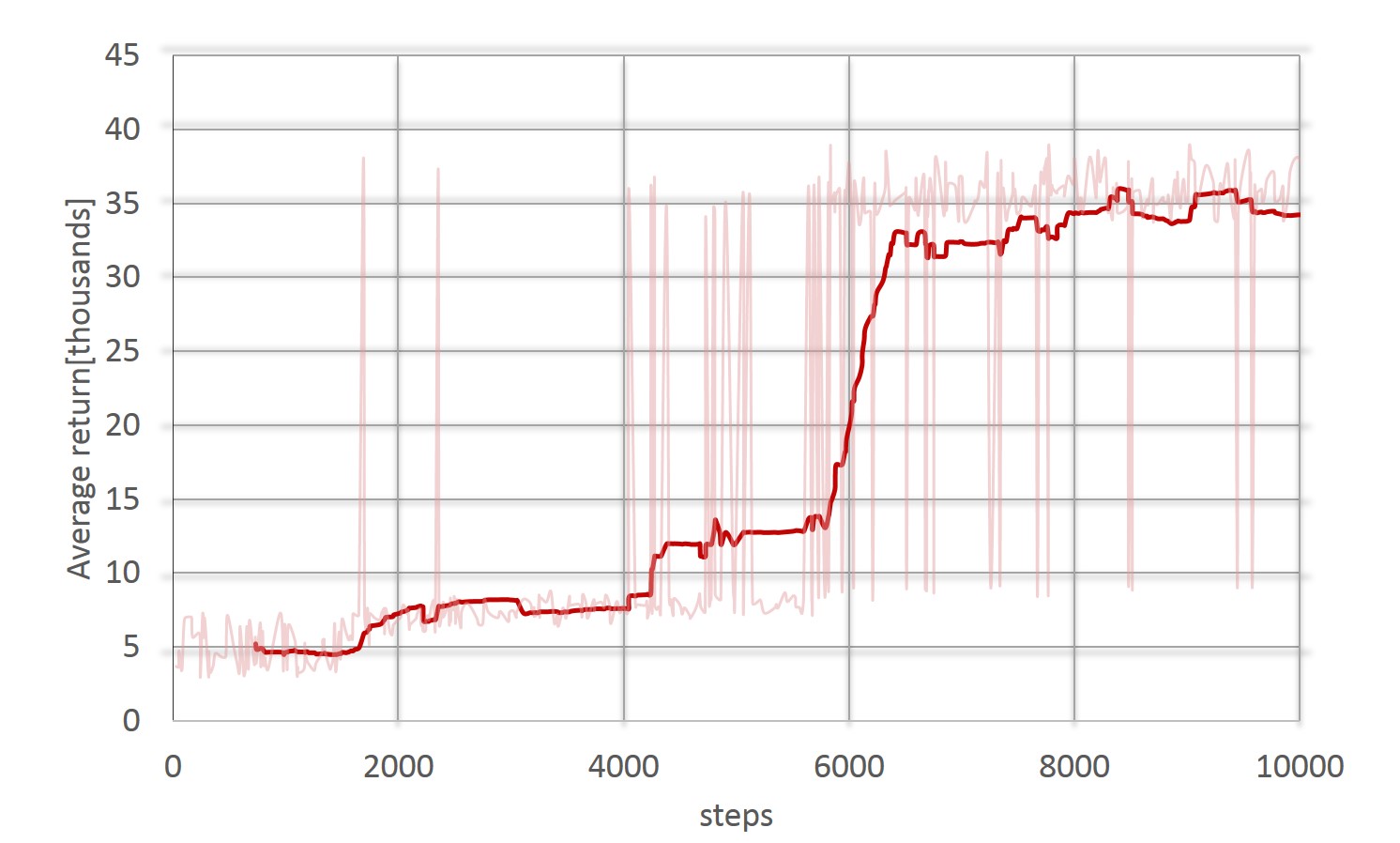}
	\caption{Training curve: average return over  $100$ steps (bold) and trial-to-trial reward (light) as a function of steps.}
	\label{fig:episodic_reward}
\end{figure}

Initial training was limited to cylinder and cuboid pegs, and was conducted with the parameters and uncertainties specified in Section \ref{TaskSU}. The training curve depicted in Fig. \ref{fig:episodic_reward} presents trial-to-trial reward  in light red and  the average reward, over $100$ steps, in bold. Improvement is slow in the initial $6000$ steps, but accelerates very quickly over the next $2000$ steps.  Trial-to-trial return fluctuates due to trial-to-trial changes in the linear and angular errors, which were selected randomly for each trial. In the initial steps, rare trials  were successful and resulted in large returns but the average return was low. As training progressed, the average return increased, and single trails with low return became more and more rare. After reaching above $35000$ points, the average reward remains  above  $33000$ and reaches above $34000$ ($92\%$ of the maximum of $38000$) at the end of training. The final policy is referred to as the nominal policy and its performance is evaluated below.

\subsection{Nominal performance}
The  performance of the nominal policy was  evaluated in simulated trials over $350$ ranges of linear and angular errors. Mean  success rate and time to complete the task in each range are detailed in  Fig. \ref{fig:original_peg_in_hole_success} and Fig. \ref{fig:original_peg_in_hole_time}. 
A trial was considered successful if the tip of the peg passed through the hole and reached the table within $\epsilon= 20[mm]$ in the allocated time of $60$[s]. Reported results at each error-range were evaluated over  $N_{test}=200$ trials.  Each trial was initiated with  randomly selected initial arm configuration and board location with uniformly distributed errors within $\pm$ the indicated linear and angular error-range. Thus,  while  success rates tend to degrade as errors increase, actual variations from one error-range to the next may not follow this trend due to random variations in selected errors and initial conditions.

Further evaluation and comparison is facilitated by dividing the error-plane into quarters as marked by black lines  in Fig. \ref{fig:original_peg_in_hole_success}.  Table \ref{tab:summary results} in the Appendix summarizes the statistics of the success rate per quarter, in terms of  mean, maximum and minimum. A maximum success rate of $99.5\%$ was achieved in the first quarter and a minimum of $77.5\%$ in the fourth quarter.  Hypothesis testing was performed for each pair of quarters, testing the null hypothesis that the samples of the success rate are from the same distribution, against the alternative hypothesis that the mean success rate increase  in the indicated order. Here and in the rest of the paper, hypothesis testing was conducted using one-sided Mann–Whitney U test with significance level $\alpha=0.05$. As reported in the Table,  success rates degrade significantly from one quarter to another, with the largest mean in the first quarter and the lowest mean in the fourth quarter. 

Fig. \ref{fig:original_peg_in_hole_time} indicates that the mean time to complete the task tends to increase gradually as linear and angular errors increase. As noted before, actual variation from one error-range to the next may not follow the trend due to random variations in selected errors and initial conditions.  Table \ref{tab:summary results time} in the Appendix summarizes the statistics of the time to complete the task in each quarter of the error-plane. A  minimum of $1.46$[s] was achieved in the first quarter and a maximum of  $7.23$[s] in the fourth quarter.  Hypothesis testing was conducted for each pair of quarters, testing  the null hypothesis that the samples of the mean time to complete the task are from the same distribution, against the alternative hypotheses that the mean time increases  in the indicated order.  As reported in the table,  mean times to complete the task increase significantly from quarter to quarter.

The results indicate that performance, in terms of both success rate and time to complete the task,  depends strongly on the accuracy of the visual system and the variations in picking the peg. Visual accuracy determines the error in horizontal location $x, y$, and orientation $\theta_z$ of the board, while the way the peg is picked determines its initial height $z$ and orientation $\theta_x, \theta_y$, so together they define the linear and angular errors as detailed in Section \ref{TaskSU}. 
 
\begin{center}
	\includegraphics[height=12cm,width=16cm]{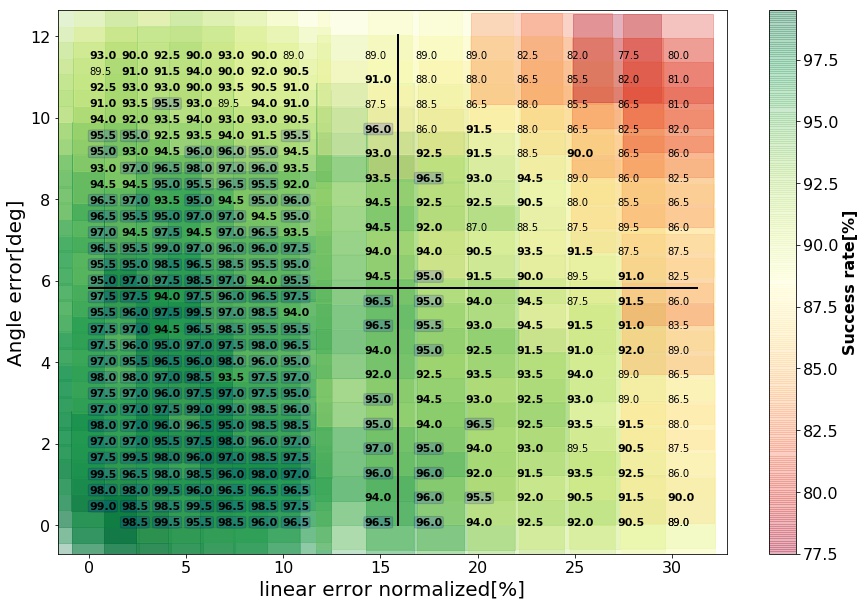}
	\centering
	\captionof{figure}{Success rates achieved by the nominal policy as a function of the range of  angular and linear errors under the same experimental conditions as those  used for training. Each success rate was estimated from $N_{test}=200$ trials with randomly selected arm configurations, board locations and uniformly distributed errors within $\pm$ the indicated linear and agular error-ranges. Values above $95\%$ are \colorbox{cyan!20}{\textbf{squared and bolded}}. Values above $90\%$ are \color{black} \textbf{bolded}.}
	\label{fig:original_peg_in_hole_success}
\end{center}
\newpage
\begin{center}
	\includegraphics[height=12cm,width=16cm]{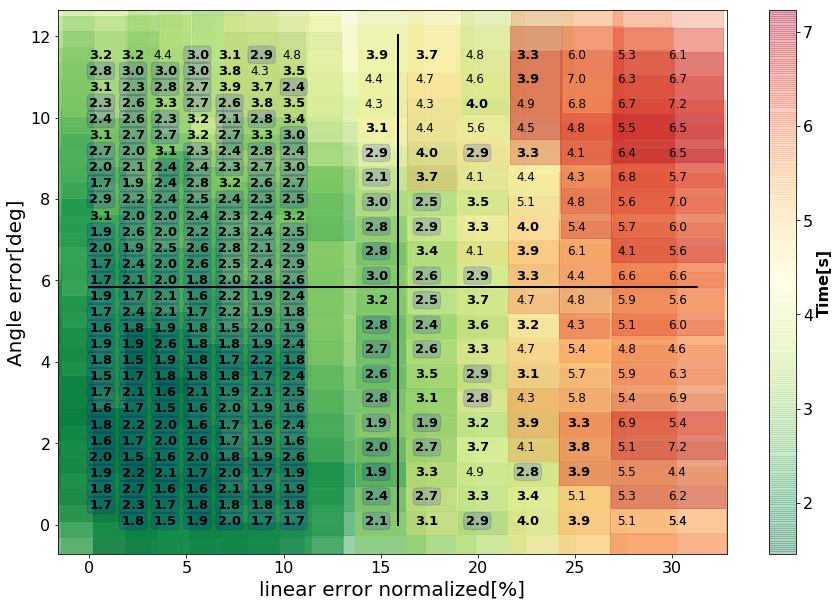}
	\centering
	\captionof{figure}{Mean time to complete the task with the nominal policy. See Fig. \ref{fig:original_peg_in_hole_success} for more details. Values below $3$[s] are \colorbox{cyan!20}{\textbf{squared and bolded}}. Values below $4$[s] are \color{black} \textbf{bolded}.}
	\label{fig:original_peg_in_hole_time}
\end{center}


\subsection{Generalization over different sizes}
As mentioned in Section \ref{TaskSU}, the nominal policy was trained  on cylindrical and cuboid pegs with characteristic size $S=50$[mm]. Performance with larger pegs ($S=60$[mm]) was evaluated with $N_{test}=50$ trials at each of $100$ error-ranges. Success rates, shown in Fig. \ref{fig:larger_peg_in_hole_success},  demonstrate good generalization.  Generalization was further assessed by comparing the  success rates with $S=60$[mm] to those obtained with $S=50$[mm] in each quarter of the error-plane (Mann–Whitney U test, $\alpha=0.05$). Table  \ref{tab:summary results} indicates that in the first and third quarters there are no significant differences between the success rates with $S=60$[mm] and $S=50$[mm]. While the characteristic size had a significant effect on the success rates in the second and fourth quarters, the differences in the mean success rates are less than $5\%$.

Table \ref{tab:summary results time} indicates that the mean time to complete the task increases by $0.65$[s] in the first quarter and by less than $1.8$[s] in the fourth quarter.  The increase in time can be attributed to  differences in magnitude of interaction forces and moments due to the larger peg size. 

 
Generalization to smaller pegs and holes is more challenging since tracking errors are relatively large. Fig \ref{fig:25mm peg in a hole} and Table \ref{tab:summary results} demonstrate good generalization in terms of success rates when inserting pegs with characteristic size $S=25$[mm] into correspondingly smaller holes.  While success rates are significantly lower in the first and second quarters, the changes in  mean success rates are less than $3\%$. Interestingly, success rates in the third quarter, with large linear errors and small angular errors, are better than the success rates in the second quarter, with small linear errors and large angular errors.  

Table \ref{tab:summary results time} indicates that the mean time to complete the task increases by less than $1$[s] in the first quarter and by less than $1.8$[s] in the second quarter, and is even shorter in the third and fourth quarters, compared to the nominal performance. Performance, in terms of both the success rate and time to complete that task, is not statistically different between the first and third quarters or  the second and fourth quarters. This may be attributed to the large  magnitude of tracking errors compared to imposed linear errors, so performance depends mainly on angular errors.

\begin{center}
	\includegraphics[height=9cm,width=14cm]{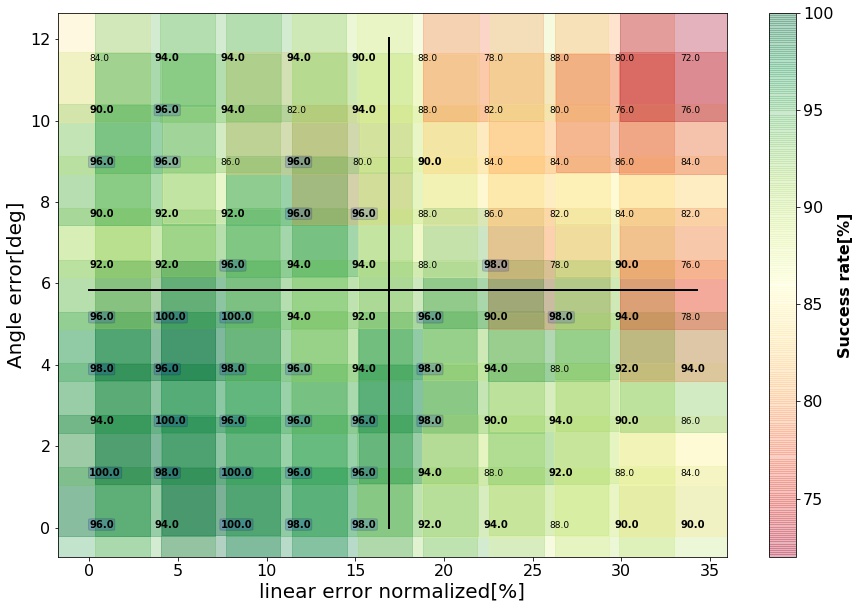}
	\centering
	\captionof{figure}{Success rates achieved by the nominal policy when inserting larger pegs  ($S=60$[mm]) in correspondingly larger holes estimated from $N_{test}=50$ trials. See Fig. \ref{fig:original_peg_in_hole_success} for more details. Values above 95$\%$ are \colorbox{cyan!20}{\textbf{squared and bolded}}. Values above 90 are \color{black} \textbf{bolded}.}
	\label{fig:larger_peg_in_hole_success}
\end{center}

\begin{center}
	\includegraphics[height=9cm,width=14cm]{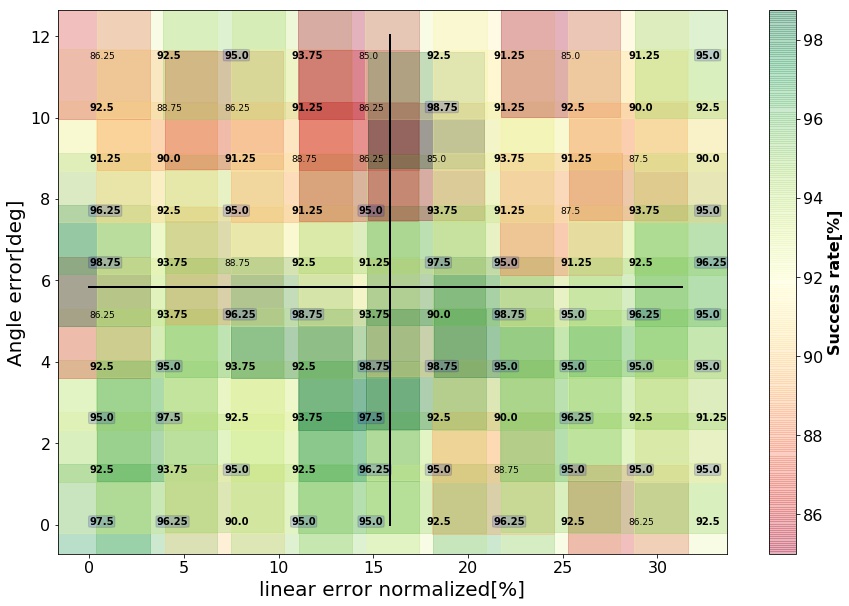}
	\centering
	\captionof{figure}{Success rates achieved by the nominal policy when inserting smaller pegs  ($S=25$[mm]) in correspondingly smaller holes estimated from $N_{test}=50$ trials. See Fig. \ref{fig:original_peg_in_hole_success} for more details. Values above 95$\%$ are \colorbox{cyan!20}{\textbf{squared and bolded}}. Values above 90 are \color{black} \textbf{bolded}.}
	\label{fig:25mm peg in a hole}
\end{center}


\subsection{Generalization over closely-related scenarios}
As mentioned in Section \ref{TaskSU}, the nominal policy was trained with cylindrical and cuboid pegs. Generalization over closely-related scenarios was evaluated by testing the performance of the nominal policy on pegs with triangular cross sections. Success rates were evaluated on $N_{test}=50$ trials at each of $100$  error-ranges. Figure \ref{fig:triangle peg in a hole} and Table \ref{tab:summary results} demonstrate that while degradation in performance is significant, the  success rates in the first quarter of the error-plane remain above $66\%$ with average of  $75\%$. These results indicate  that the nominal policy is not over fitted to cylindrical and cuboid pegs, and suggest that the nominal policy may provide a good baseline policy for learning to perform peg-in-hole using other pegs, as evaluated next.
\newpage
\begin{center}
	\includegraphics[height=10cm,width=16cm]{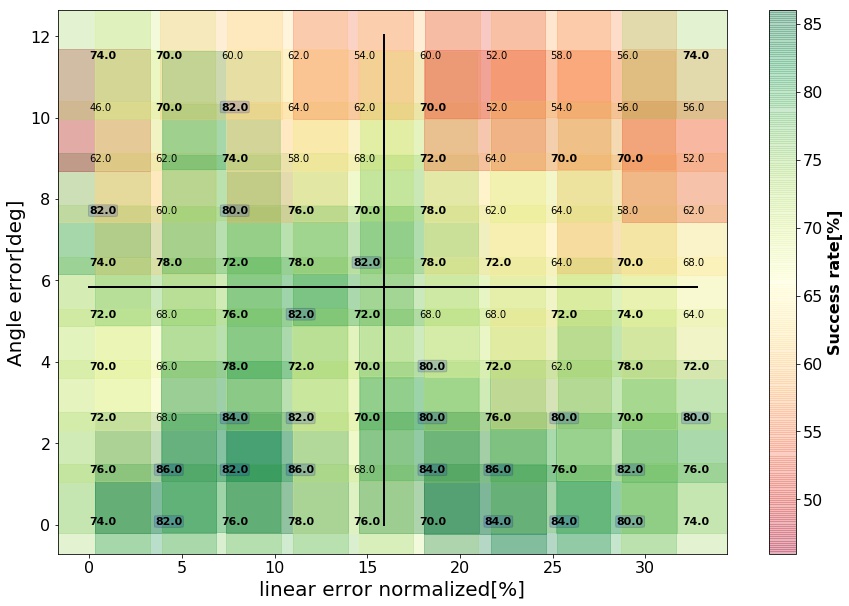}
	\centering
	\captionof{figure}{Success rates achieved by the nominal policy when inserting  pegs with triangular cross-sections, which were not used during original training. Each success rate was estimated from $N_{test}=50$ trials. See Fig. \ref{fig:original_peg_in_hole_success} for more details. Values above $80\%$ are \colorbox{cyan!20}{\textbf{squared and bolded}}. Values above $70\%$ are \color{black} \textbf{bolded}.}
	\label{fig:triangle peg in a hole}
\end{center}

\subsection{Transfer learning}
Generalization to new shapes (investigated in the previous sub-section) evaluates how well a nominal policy, which was trained on pegs with specific shapes (here cylindrical and cuboid pegs), may generalize to pegs with another shape (here pegs with triangular cross-sections) without any further training. Transfer learning, investigated here, evaluates how fast and how well the nominal policy can be re-trained to perform the task with pegs of another shape. Starting with the nominal policy, Fig. \ref{fig:Triangle learning} depicts the average return during subsequent training using pegs with triangular cross-sections. It is evident that only 800 trials were required to re-optimize the nominal policy for inserting pegs with triangular cross-sections and improve the average return from  around $24000$ to around $34000$. This is a reduction of an order of magnitude in the number of trials,  compared to initial learning (Fig. \ref{fig:episodic_reward}).

Fig. \ref{fig:triangle after learning} and Table \ref{tab:summary results} depict the resulting improvement in success rate for inserting  pegs with triangular cross-sections. Success rates were evaluated on $N_{test}=50$ trials at each of $100$  error-ranges.  The success rates  after re-training are significantly higher ($\alpha=0.05$)  than those before re-training in all  quarters of the error-plane. Furthermore, the success rates in the first quarter of the error-plane cannot be statistically distinguished from the nominal success rates.  While the success rates in the other quarters of the error-plane are  significantly lower than the nominal success rates, the differences in the means are less than $7,6,11\%$ in the second, third and fourth quarters, respectively. The lower performance may be attributed to the complexity of the task.

Table \ref{tab:summary results time} indicates that the time to complete the task in the first and third quarters of the error-plane are even shorter than the nominal times. The time to complete the task in the second and fourth quarters of the error-plane are statistically indistinguishable from nominal.

\begin{center}
	\includegraphics[height=8cm,width=12cm]{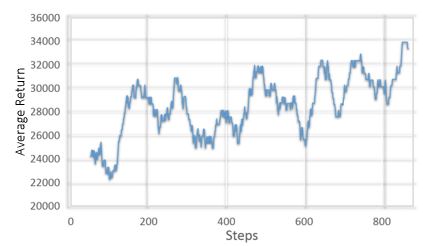}
	\centering
	\captionof{figure}{Re-training curve: average return over $100$ steps as a function of steps, while re-training the nominal policy on pegs with triangular cross-section, which were not used during original training.}
	\label{fig:Triangle learning}
\end{center}
\newpage
\begin{center}
	\includegraphics[height=9cm,width=14cm]{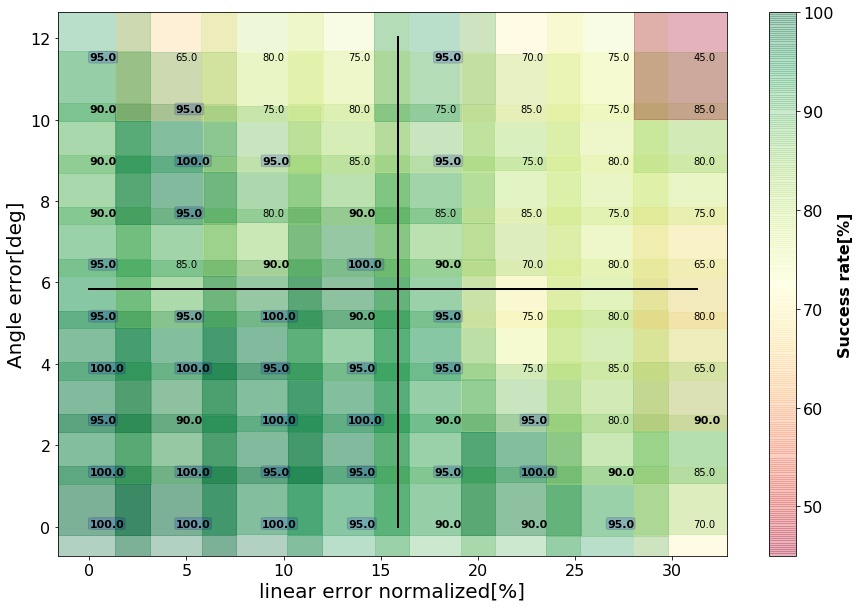}
	\centering
	\captionof{figure}{Success rates achieved by the re-trained policy when inserting  pegs with triangular cross-section on which the policy was re-trained. Each success rate was estimated from $N_{test}=50$ trials. See Fig. \ref{fig:original_peg_in_hole_success} for more details. Values above $95\%$ are \colorbox{cyan!20}{\textbf{squared and bolded}}. Values above $90\%$ are \color{black} \textbf{bolded}.}
\label{fig:triangle after learning}
\end{center}


\subsection{Vision to action}
In the above sub-sections, the nominal policy was evaluated with respect to a range of imposed uncertainties reflecting errors in visual localization and grasping. To validate the range of imposed uncertainties, and the performance of the combined vision to action, we evaluated the nominal policy when extracting object localization from synthetic images.  Images ($1024 \times 1024 \times 3$) were taken by a simple 2D 1Mpx camera in MuJoCo. Bird-view images were generated since they are useful for planning consecutive actions. A YOLO-like model was trained to localize the holes in the board (see Fig. \ref{fig:board_location}) within a maximum error of $8$[mm] in $x$ and $y$ and $8^\circ$ in planar rotation. The height above the table (i.e., the location along the z axis) was assumed to be known. However, uncertainties in the location and orientation at which the peg was grasped were accounted for by setting the orientation errors around $x$ and $y$ and location errors along the $z$-axis to their maximum levels ($12^\circ$ and $30\%$, respectively).  

Combining the nominal policy with object localization from the visual system resulted in $92\%$ success rate over $N_{test}=200$ trials for both cylindrical and cuboid pegs. This result fits well with the nominal performance, being slightly above the mean success rate in the third quarter of the error-space ($91.8\%$ for Q3 in Table \ref*{tab:summary results})  and slightly less than  the mean success rate in the second quarter ($94.2\%$ for Q2 in Table \ref*{tab:summary results}).  Better results can be achieved by minimizing grasping errors or by training a YOLO model to achieve smaller localization errors.    

\section{Simulation-to-real transfer}
\label{s:Robot}
The final experiment evaluates how well the policy learned in simulation can be transferred to a real robot without further training. The experiment was conducted with UR5e equipped with OnRobots HEX-E force sensor that measures the forces and moments at the EEF. To accurately capture the contact forces and to generate smooth movements, high frequency communication was used (Real Time Data Exchange). Specifically, force and position measurements were sampled at 500Hz and effort commands were sent at 125Hz. 
  
Since our focus is on evaluating  the control method rather than localization accuracy, we added stochastic localization errors to the robot's goal in (x,y) plane and in the orientation around the z-axis, as described in Section \ref{TaskSU}. Furthermore, no effort was made to grasp the peg in a precise height or orientation. Table \ref{tab:real robot results} summaries the experimental results with UR5e based on 10 trials for six different pegs (rows), demonstrating generalization over shape and size, and different ranges of localization uncertainties (columns).  These results indicate that the policy trained using the proposed method can transfer seamlessly to real world. This can be attributed to two strategies inherent in the proposed method: (i) working in Cartesian space of the EEF, while relying on the robot's controller to make the conversion to joint space, and (ii) learning compliance, which  is inherently robust to uncertainties including those associated with the transition to real world.

\begin{table}[!htb]
	\centering
	\caption{Success rate with a UR5e robot over different board locations, peg-shapes and sizes under different ranges of linear and angular errors}
	\label{tab:real robot results}
	\begin{tabular}{|l|c|c|c|c|c|}
		\hline
		\rowcolor[HTML]{DAE8FC} 
		Peg & \multicolumn{5}{c|}{\textbf{Error-range linear / angular}}                                         \\ \hline
		Cross-section (size) & 3mm/12deg & 4mm/12deg & 6mm/12deg & 8mm/12deg & 10mm/12deg \\ \hline
		Cylindrical (50mm) &           &           & 10/10     & 10/10     & 8/10       \\ \hline
		Cubic (50mm)     &           &           & 10/10     & 9/10      & 9/10       \\ \hline
		Triangular (50mm) &         &           & 10/10     & 7/10      & 5/10       \\ \hline
		Cylindrical (25mm) & 10/10     & 10/10     & 10/10     &           &            \\ \hline
		Cubic (25mm)     & 10/10     & 10/10     & 9/10      &           &            \\ \hline
		Triangular (25mm) & 10/10     & 8/10      &           &           &            \\ \hline
	\end{tabular}
\end{table}

\section{Conclusions}
\label{conclusion}

The reported research extends the framework of dynamic movement primitives (DMP) to manage contact-rich assembly tasks. We introduce a coupling term that enforces the desired compliance in response to interaction forces. The coupling term adapts the pre-learned trajectory to task variations in response to the resulting interaction forces. 

DMP was adopted to facilitate working in Cartesian space and dividing the task to movement primitives. 
Working in  Cartesian space, rather than joint space, is critical for contact-rich tasks, since  interaction forces are related to joint torques via the Jacobian of the robot, which depends non-linearly on the joint angles,  as explained in Section \ref{ss:AS}. Dividing the task into different movement primitives, including free and contact-rich movements,  is important to facilitate learning different skills. Individual skills require smaller sets of parameters thereby resulting in dimensional reduction.   Further pruning of the search space is achieved by learning the parameters of the desired compliance rather than the mapping from forces to control signals. These strategies enhance sample efficiency, so it was possible to take advantage of model-free on-policy RL algorithms, and in particular PPO, which  guarantees convergence and reduces brittleness.  

Our results demonstrate that pruning the search space facilitates efficient learning of insertion skills. As mentioned in Section \ref{s:INTRO}, skills should generalize across space, size, shape and  closely related scenarios while handling large uncertainties. As summarized in Table \ref*{tab:summary results}, the proposed method is indeed: (1) invariant to hole location (Fig. \ref{fig:original_peg_in_hole_success}), (2) invariant to hole size (Fig.  \ref{fig:25mm peg in a hole} and Fig. \ref{fig:larger_peg_in_hole_success}), (3) invariant to hole shape (Fig. \ref{fig:original_peg_in_hole_success}), (4) transferable to closely related scenarios, and, in particular, to pegs with previously unseen cross-sections (Fig. \ref{fig:triangle after learning}), and (5) robust to large uncertainties (Section \ref{TaskSU}).

The nominal policy was evaluated over $350$ different ranges of linear and angular errors. Success rates decreased from $99.5\%$  to $77.5\%$ as the range of linear and angular errors increased to $8$[mm] and $12^\circ$, respectively, while the time to complete the task increased from $1.46$[s] to $7.23$[s] (Fig. \ref{fig:original_peg_in_hole_success} and Fig. \ref{fig:original_peg_in_hole_time}). Linear and angular errors were included to account for localization and grasping uncertainties. The expected accuracy of visual localization, and the performance of the combined vision to action, was evaluated by extracting object localization from synthetic images. A YOLO model was trained to achieve object localization with errors below $8$[mm] and $8^\circ$. Combining the resulting localization with the nominal policy resulted in $92\%$ success rate for both cylindrical and cuboid pegs. Better results can be achieved by minimizing grasping errors or by training a YOLO model to achieve smaller localization errors.  

Finally, we demonstrated that the proposed method overcomes the sim-to-real challenge. As detailed in Section \ref{s:Robot}, the policy learned in simulations achieved high success rates on a real robot (UR5e), inserting pegs of different shapes and sizes. These results indicate that the policy trained using the proposed method can transfer seamlessly to real world. This can be attributed to two strategies inherent in the proposed method (i) working in Cartesian space of the EEF, and (ii) learning compliance, which  is inherently robust to uncertainties. 

The proposed method is unique in the range of uncertainties and variations that it can handle. For example, the tilt strategy  (\cite{Chhatpar2001}), sensory-guided search (\cite{Newman2001}), spiral search combined with a contact identification method  (\cite{Jasim2014}), and Gaussian Mixture Model (GMM) derived from human demonstration (\cite{tang2015learning}), were demonstrated to achieve high success rate on cylindrical pegs, but did not demonstrate generalization over space, size, shape or closely related scenarios. More detailed comparison is hampered due to task differences and lack of statistically significant information.

The proposed method is a promising basis for learning other skills for contact-rich tasks, including tasks involving flexible materials, such as wiring. The proposed method provides also the basis for learning sequences of operations to complete contact-rich tasks.

\section*{Acknowledgment} We would like to acknowledge support for this project from  Gordon Center for System Engineering (grant $\#$ 2026779) and Israel Ministry of Science and Technology (grant $\#$2028449). 


\newpage
\bibliography{sample}
\newpage
\appendix
\section{Statistical results}
\begin{table}[ht]
	\centering
	\caption{Success rates     summarized per quarter for different conditions. Hypothesis testing (HT) was conducted using Mann–Whitney U test with significance level $\alpha=0.05$. The null  hypothesis in each case was that the relevant distributions of success rates are indistinguishable. HT across quarters was performed with respect to the alternative hypothesis that the mean  success rates in the different quarters are ordered as indicated. HT compared to nominal performance (perf.) was performed with respect to the alternative hypothesis that the mean success rate in each quarter is smaller than the nominal performance. HT comparing success rates before and after re-training was performed with respect to the alternative hypothesis that the mean success rate in each quarter increased by re-training.}
	\label{tab:summary results}
	\begin{tabular}{||c ||c ||c ||c || c|| }
		\hline
		\rowcolor[HTML]{DAE7FC} 
		-& \textbf{\begin{tabular}[c]{@{}c@{}}Q1-\\bottom left \end{tabular}} & \textbf{\begin{tabular}[c]{@{}c@{}}Q2-\\upper left \end{tabular}} & \textbf{\begin{tabular}[c]{@{}c@{}}Q3-\\bottom right \end{tabular}} & \textbf{\begin{tabular}[c]{@{}c@{}}Q4-\\upper right \end{tabular}} \\ \hline
		& \multicolumn{4}{c|}{\textbf{Section 6.2-Nominal performance}}                                         \\ \hline
		\textbf{mean} & 96.96                   & 94.22                  & 91.83                    & 87.83                   \\ \hline
		\textbf{min}  & 92                      & 87.5                   & 83.5                     & 77.5                    \\ \hline
		\textbf{max}  & 99.5                    & 99.0                   & 96.5                     & 96.5                    \\ \hline
		\textbf{\begin{tabular}[c]{@{}c@{}}Quarters' order\end{tabular}}  & $>$2,3,4                   &  $>$3,4                  &  $>$4                     &                     \\ \hline
		& \multicolumn{4}{c|}{\textbf{Section 6.3-Generalization to different sizes}}                            \\ \hline
		& \multicolumn{4}{c|}{{\color[HTML]{656565} \textbf{60mm}}}                                             \\ \hline
		\textbf{mean} & 96.88                   & 92                     & 91.2                     & 83.52                   \\ \hline
		\textbf{min}  & 92                      & 80                     & 78                       & 72                      \\ \hline
		\textbf{max}  & 100                     & 96                     & 98                       & 98                      \\ \hline
		\textbf{\begin{tabular}[c]{@{}c@{}}Quarters' order\end{tabular}}  & $>$2,3,4                   &  $>$4                  &  $>$4                     &                    \\ \hline
		& \multicolumn{4}{c|}{{\color[HTML]{656565} \textbf{25mm}}}                                             \\ \hline
		\textbf{mean} & 94.45                    & 91.5                     & 93.8                     & 92.05                    \\ \hline
		\textbf{min}  & 86.25                      & 85                     & 86.25                       & 85                      \\ \hline
		\textbf{max}  & 98.75                     & 98.75                    & 98.75                      & 98.75                     \\ \hline
		\textbf{\begin{tabular}[c]{@{}c@{}}Quarters' order\end{tabular}}  & $>$2,4                  &                & $>$2,4                     &                    \\ \hline
		& \multicolumn{4}{c|}{\textbf{Section 6.4-Generalization to different shape}}                           \\ \hline
		\textbf{mean} & 75.44                   & 68.8                   & 75.68                    & 63.68                   \\ \hline
		\textbf{min}  & 66                      & 46                     & 62                       & 52                      \\ \hline
		\textbf{max}  & 86                      & 82                     & 86                       & 78                      \\ \hline
		\textbf{\begin{tabular}[c]{@{}c@{}}HT: Nominal pref. \end{tabular}}  &True                   & True                     & True                       & True   \\ \hline
		& \multicolumn{4}{c|}{\textbf{Section 6.5-Transfer learning}}                                           \\ \hline
		\textbf{mean} & 97                      & 87.5                   & 86                       & 78                      \\ \hline
		\textbf{min}  & 90                      & 65                     & 65                       & 45                      \\ \hline
		\textbf{max}  & 100                     & 100                    & 100                      & 95                      \\ \hline
		\textbf{\begin{tabular}[c]{@{}c@{}}Quarters' order\end{tabular}}  & $>$1,2,3,4                   & $>$4                  & $>$4                  &                   \\ \hline
		\textbf{\begin{tabular}[c]{@{}c@{}}HT: Nominal pref. \end{tabular}}  &False                    & True                     & True                       & True   \\ \hline
		\textbf{\begin{tabular}[c]{@{}c@{}}{HT: after vs.} \\ {before re-training} \end{tabular}}  &True                    & True                     & True                       & True   \\ \hline
	\end{tabular}
\end{table}
\newpage
\begin{table}[ht]
	\centering
	\caption{Mean time to complete the task summarized per quarter for different condition. Hypothesis testing (HT) was conducted using Mann–Whitney U test with significance level $\alpha=0.05$. The null  hypothesis in each case was that the relevant distributions of mean times are indistinguishable. HT across quarters was performed with respect to the alternative hypothesis that the mean of the distributions of mean times in each quarter is ordered as indicated. }
	\label{tab:summary results time}
	\begin{tabular}{||c ||c ||c ||c || c||}
		\hline
		\rowcolor[HTML]{DAE8FC} 
		-& \textbf{\begin{tabular}[c]{@{}c@{}}Q1-\\bottom left \end{tabular}} & \textbf{\begin{tabular}[c]{@{}c@{}}Q2-\\upper left \end{tabular}} & \textbf{\begin{tabular}[c]{@{}c@{}}Q3-\\bottom right \end{tabular}} & \textbf{\begin{tabular}[c]{@{}c@{}}Q4-\\upper right \end{tabular}} \\ \hline
		& \multicolumn{4}{c|}{\textbf{Section 6.2-Nominal performance}}                                         \\ \hline
		\textbf{mean} & 1.9                  & 2.72                  &4.34                    & 4.88                   \\ \hline
		\textbf{min}  & 1.46                      &1.69                   &1.94                    & 2.48                   \\ \hline
		\textbf{max}  & 3.16                   & 4.76                  & 7.17                     & 7.23                    \\ \hline
			\textbf{\begin{tabular}[c]{@{}c@{}}Quarters' order\end{tabular}}  & $<$2,3,4                   &  $<$3,4                  &  $<$4                     &                     \\ \hline
		& \multicolumn{4}{c|}{\textbf{Section 6.3-Generalization to different sizes}}                            \\ \hline
		& \multicolumn{4}{c|}{{\color[HTML]{656565} \textbf{60mm}}}                                             \\ \hline
		\textbf{mean} & 2.57                   & 3.01                     & 5.92                     & 6.67                  \\ \hline
		\textbf{min}  & 1.1                     & 1.37                     & 2.5                       & 2.6                      \\ \hline
		\textbf{max}  & 5.11                     & 6.13                     & 9.72                       & 12.97                      \\ \hline
		\textbf{\begin{tabular}[c]{@{}c@{}}Hypothesis testing \\quarters\end{tabular}}  & $<$3,4      & $<$3,4     &        &  \\ \hline	
		& \multicolumn{4}{c|}{{\color[HTML]{656565} \textbf{25mm}}}                                             \\ \hline
		\textbf{mean} & 2.87                   & 4.47                     & 3.09                    & 4.42                    \\ \hline
		\textbf{min}  & 0.57                     & 1                     & 1.16                       &2.33                      \\ \hline
		\textbf{max}  & 5.54                     & 8.32                    & 6.29                      & 7.08                     \\ \hline
		\textbf{\begin{tabular}[c]{@{}c@{}}Hypothesis testing \\quarters\end{tabular}}  & $<$2,4        &                  & $<$2,4       &   \\ \hline
		& \multicolumn{4}{c|}{\textbf{Section 6.5-Transfer learning}}                                           \\ \hline
		\textbf{mean} & 1.3                      & 2.87                   & 3.12                       & 4.96                      \\ \hline
		\textbf{min}  & 0.7                      & 0.93                     & 0.8                       &0.98                      \\ \hline
		\textbf{max}  & 2.94                     & 5.15                    & 6.09                      & 10.33                      \\ \hline
		\textbf{\begin{tabular}[c]{@{}c@{}}Hypothesis testing \\quarters\end{tabular}}  & $<$2,3,4       &   $<$4                 &   $<$4       &    \\ \hline	
	\end{tabular}
	
\end{table}



\end{document}